\documentclass{article}

\usepackage[preprint]{neurips_2026}

\workshoptitle{Foundation Models for the Brain and Body}

\usepackage[utf8]{inputenc} 
\usepackage[T1]{fontenc}    
\usepackage{hyperref}       
\usepackage{url}            
\usepackage{booktabs}       
\usepackage{amsfonts}       
\usepackage{nicefrac}       
\usepackage[expansion=false]{microtype}      
\usepackage{xcolor}         
\usepackage{graphicx}
\usepackage{tabularx}
\usepackage{pifont}
\usepackage{comment}
\usepackage{array}
\usepackage{amsmath}
\title{Representational and Functional Robustness to Electrode Montages in EEG Foundation Models}

\author{%
  Jakob Steglich$^{1}$ \hspace{8pt}
  Justus Meyer zu Bexten$^{1,2}$ \hspace{8pt}
  Shakiba Moradi$^{3,4}$ \hspace{8pt}
  Laure Ciernik$^{4,5}$ \And
  Simon M. Hofmann$^{1,*}$ \hspace{8pt}
  Mina Jamshidi Idaji$^{3,4,*}$ \\[12pt]
  \normalsize
  $^1$ Max Planck Institute for Human Cognitive and Brain Sciences, Leipzig, Germany \\
  $^2$ ScaDS.AI Dresden/Leipzig \\
  $^3$ BIFOLD--Berlin Institute for the Foundations of Learning and Data, Berlin, Germany \\
  $^4$ Machine Learning Group, Technische Universität Berlin, Berlin, Germany \\
  $^5$ Hector Fellow Academy, Karlsruhe, Germany \\
  $^*$Equal supervision contribution\\
  Correspondence to: mina.jamshidi.idaji@tu-berlin.de}

\begin{document}

\maketitle

\begin{abstract}
EEG foundation models (EEG-FMs) are intended to generalize across different datasets by learning representations that, ideally, are invariant to dataset-specific EEG configurations such as electrode montages. 
However, EEG-FMs that accept different montages as input do not guarantee that representations and predictions remain stable across different electrode configurations, especially outside the training setting.
In this work, we investigate the effects of different electrode montages through a joint functional and representational analysis of four EEG foundation models selected to span distinct montage-handling designs. 
We evaluate embeddings on cross-subject resting-state eyes-open/closed and within-subject motor-imagery classification under spatially informed channel reduction. 
Functional robustness is tested through the generalizability of linear probes across channel counts, while representational robustness is assessed through within-subject similarity and preservation of between-subject geometry.
The four models show distinct robustness profiles, and the two axes dissociate: large changes in embedding similarity need not come with comparable probe degradation, and stable embeddings can still lose downstream performance. Comparing two readouts of the same encoder further shows that aggregation, not the encoder alone, determines functional robustness: pooling into anatomically aligned regions degrades less than a learned global readout, despite being montage-invariant by construction.
Montage robustness is therefore a joint property of the encoder and its aggregation, and characterizing it requires both a representational and a functional axis. Input compatibility alone is evidence for neither.
\end{abstract}

\section{Introduction} \label{intro}

Electroencephalography (EEG) datasets vary not only in their objectives and participants, but also in their hardware setup, such as electrode layout (montage).
Computational and statistical models of EEG are usually tuned to a specific dataset and rarely transfer to other studies and applications, particularly to out-of-lab deployments where electrode count and position differ from the recording setup.
Recent large, pretrained EEG foundation models (EEG-FMs) leverage deep learning methods on heterogeneous datasets with the promise to overcome this lack of transferability \citep{aristimunhaEEGFoundationChallenge2025, xiongEEGFMBenchComprehensiveBenchmark2026}, ideally extracting representational spaces from diverse data that are channel- and task-invariant.

Existing EEG-FMs follow various strategies to accommodate different electrode montages.
Channels are modeled independently \citep{sukhbaatarSingLEMSingleChannelLarge2025a}, or together through explicit spatio-temporal attention \citep{wangCBraModCrissCrossBrain2024a}.
Some approaches unify them in fixed-size latent representations \citep{donerLUNAEfficientTopologyAgnostic2025, ouahidiREVEFoundationModel2025}, using Perceiver-style \citep{jaegle2021perceiver} cross-attention (structured overview in ~\autoref{app:sec:EEG_FM_overview}).
Despite efforts to accommodate various channel setups in the input domain of EEG-FMs, their ability to extract representations invariant to electrode configurations remains largely unexplored. 

To address this gap, we present a joint functional and representational analysis of montage robustness in four EEG-FMs selected to span distinct montage-handling designs.
We study a common setting in EEG-FM deployment, in which a downstream predictor encounters a different number of channels during inference than were available during pretraining.
We assess functional robustness (the effect on a model's output) by testing the generalizability of linear probes across channel counts and representational robustness (the effect on latent embeddings) \citep{klabundeSimilaritySurvey2025} through the preservation of within-instance and between-subject representational similarity.
We evaluate these properties on cross-subject eyes-open/closed and within-subject motor-imagery classifications.
Across these settings, representational and decision stability do not consistently coincide, indicating that both dimensions are needed to characterize overall montage robustness.


\section{Related Work}\label{rel-work}
Recent benchmarks have shown that EEG-FMs frequently fail to match specialist baselines \citep{liuEEGFMCompassProgressBenchmarking2026} and often encode dataset origin rather than transferable clinical features \citep{zareNegativeControl2026}. Their robustness to variations in electrode montages has been recently addressed through adaptation methods 
\citep{maAdaptingFrozenFoundation2026,kokateChannelAdaptation2026} and evaluation protocols based on channel masking, sparse subsampling, or lobe-restricted montages \citep{liOmniBenchFutureUniversal2025, kommineniMultidimensionalFrameworkEvaluating2026a}.
\citet{sircaBeyondAccuracy2026} perturbed six EEG-FMs with noise and random or region-based channel dropout, reporting degradation in all of them and greater tolerance to electrode omission than to zero-masking.

Whereas these studies evaluate downstream robustness or methods for adapting models to montage mismatch, our work asks whether channel-count reductions alter representations in ways that correspond to downstream functional stability (cf. \citep{klabundeSimilaritySurvey2025,dingGrounding2021}).


\section{Methods} \label{methods}
\subsection{Foundation Model Embeddings from EEG Data}\label{sec:fm_embeddings}

\textbf{Evaluated models.}
We evaluated four state-of-the-art EEG-FMs, namely
SingLEM \citep{sukhbaatarSingLEMSingleChannelLarge2025a},
CBraMod \citep{wangCBraModCrissCrossBrain2024a},
LUNA \citep{donerLUNAEfficientTopologyAgnostic2025}, and
REVE \citep{ouahidiREVEFoundationModel2025}.
These represent four categories of channel fusion strategies in EEG-FMs (\autoref{app:sec:EEG_FM_overview}), and enable us to investigate how this architectural aspect contributes to montage robustness. 

Across all four models, we denote a multi-channel EEG recording by $S \in \mathbb{R}^{C \times T}$, where $C$ is the number of EEG channels, and $T$ the number of time points. Each model segments $S$ into $P$ temporal patches of $T_p$ time points (patches may overlap, depending on the model), yielding a patched tensor $S_p \in \mathbb{R}^{C \times P \times T_p}$.
SingLEM, CBraMod, and REVE project to per-channel embeddings $E\in \mathbb{R}^{C\times P \times F}$ with $F$ being the embedding dimension, while LUNA fuses the channel dimension into $Q$ learned query tokens, yielding $E\in \mathbb{R}^{(Q\cdot P) \times F}$; we set $Q=4$ following the original implementation.
Furthermore, REVE obtained a global query token during pretraining, which can be used to reduce its embedding to $E \in \mathbb{R}^{F}$ with attention pooling; we denote this readout of REVE as \textit{REVE-A}. See \autoref{app:sec:models_embedding} for more details.

\textbf{Embedding tensors and their pooling.}
The main challenge in investigating the robustness of EEG-FMs to changes in the channel montage is to project their embeddings to a \textit{montage-agnostic} space. 
For this purpose, we introduce a region-of-interest (ROI) pooling strategy $\phi: \mathbb{R}^{C \times P \times F} \to \mathbb{R}^{R \times P \times F}$, $\phi(E)_{r,:, :} = |\mathcal{R}_r|^{-1} \sum_{c \in \mathcal{R}_r} E_{c,:, :}$ (details in \autoref{app:sec:models_embedding}), which averages feature vectors within $R$ ROIs $\mathcal{R}_r \subseteq \{1,\dots,C\}, r=1,\cdots, R$, mapping embeddings to the same spatial dimensionality regardless of channel count. We set $R=10$ and depict the ROIs in \autoref{fig:montage-roi-mapping}. 
Because LUNA and REVE-A use a built-in channel fusion module, their embedding outputs are already montage-agnostic. Therefore, they are not directly comparable to other models.
Note, \citep{donerLUNAEfficientTopologyAgnostic2025} demonstrate how LUNA's queries learn distinct spatial profiles, similar to ROIs.
We report our results both for REVE with ROI pooling (REVE-ROI), and with attention pooling (REVE-A).
For linear probing, $\phi(E)$ or $E$ is flattened to $\mathbb{R}^{RPF}$ (\textit{roi-concat-all}) or $\mathbb{R}^{CPF}$ (\textit{concat-all}), of which only the former is montage-independent.

\subsection{Datasets and preprocessing}

\textbf{Tasks.}
We include two complementary tasks in our experiments: (1) cross-subject classification of eyes closed vs. open resting-state (RS) data and (2) intra-subject, trial-level classification of hand movement-imagery (MI) data.
For (1), closing and opening eyes has a known and pronounced neural basis, including higher alpha-band activity at the occipital region; therefore, we expect a high cross-subject classification performance.
In (2), hand MI is a localized, event-related activity and has an acceptable intra-subject discriminative performance \citep{Pfurtscheller_2001}\footnote{Note, previous reviews of EEG-FMs \citep{liuEEGFMCompassProgressBenchmarking2026} and our own experiments indicate that cross-subject generalization does not yield sufficient classification performance for the MI dataset.}.
The datasets are class-balanced, and we report AUROC as performance metrics.

\textbf{Datasets.}
We used a 100-participant subset of the 129-channel Healthy Brain Network (HBN) EEG dataset \citep{shirazi2024hbneeg} for cross-subject eyes open versus closed classification.
We retrieved 4-s epochs from each of ten trials per participant.
As channel Cz was used as reference during recording, we retained only the 128 GSN-HydroCel montage in our analyses and changed the reference to common-average.
For the MI task, \cite{lee2019openbmi} provide a dataset recorded with 62 electrodes in the standard-1005 montage from 52 participants instructed to imagine left- or right-hand movement. 
We used the 4-s imagery segments of the 200 labeled imagery trials per subject (see dataset details in \autoref{app:sec:dataset_prep_details}). 
The RS data was not used for pretraining of any model, while the MI dataset was present in the SingLEM pretraining corpus.

\section{Experiments} \label{experimental-setup}

\subsection{Experiment 1: Functional cross-montage robustness}
\label{exp-functional}

\begin{figure}[t]
    \centering
    \includegraphics[width=\linewidth]{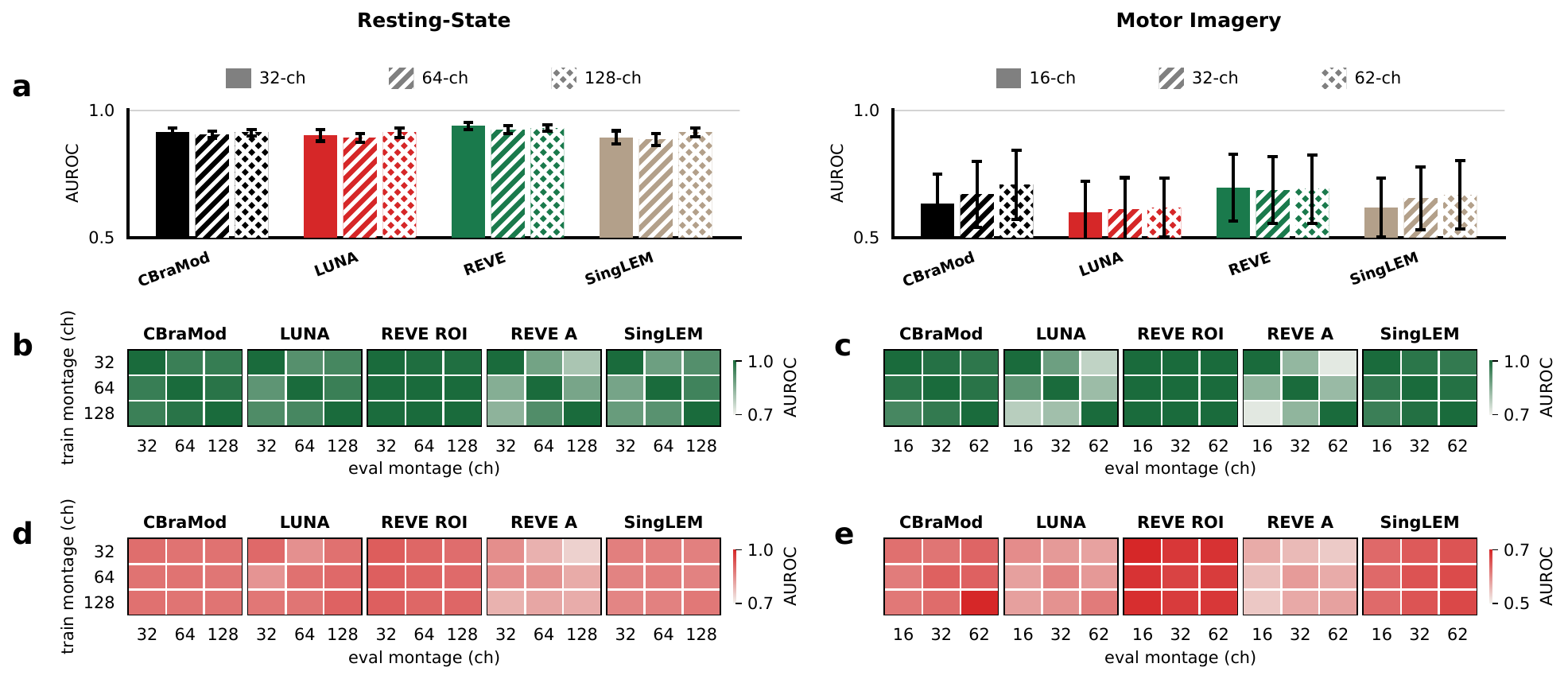}
    \caption{\textbf{Functional within- and cross-montage robustness.} \textbf{(a)} Within-montage baselines (N2N-probes with concat-all as pooling) for inter-subject RS and intra-subject MI classification; bars show mean AUROC and standard deviation across seeds (n=10) for RS and subjects for MI (\autoref{tab:N2N-prob-perf}). \textbf{(b, c)} N2M-probes on RS and MI, respectively, with matched train and test samples, and \textbf{(d, e)} with held-out test data. Rows give the training and columns the test montage. The full montage has 128 channels for RS and 62 for MI (\autoref{app:sec:dataset_prep_details}).}
    \label{fig:functional_robustness}
\end{figure}

\textbf{Protocol.} 
We trained linear probes \citep{Alain_2018}, implemented with logistic regression, on embeddings from an $N$-channel training montage and evaluated them on an $M$-channel test montage (\textbf{\textit{N2M-probes}}; $M = N$ gives the within-montage \textbf{\textit{N2N}} case). We focus on the deployment-relevant direction $M < N$, although \autoref{fig:functional_robustness} reports the full grid (see \autoref{app:sec:model_training} for details).

\textbf{Within-montage baselines.} 
To compare the raw discriminative information available in the native embeddings of each montage, we first tested N2N-probes with \textit{concat-all} as the pooling, which preserves the full embedding tensor and is therefore the least constrained of the evaluated pooling methods. We observe no clear performance differences among channel montages within any FM, indicating comparable classification-relevant information content across montages (\autoref{fig:functional_robustness}-(a, b), \autoref{tab:N2N-prob-perf}). Moreover, \textit{ROI-concat-all} consistently outperforms channel averaging on both datasets and trails \textit{concat-all} only slightly (\autoref{tab:pooling-combined}), so it retains enough task-relevant information while keeping a montage-independent dimensionality.

The following two experiments are done with ROI-concat-all as the pooling method with $R=10$, see \autoref{fig:functional_robustness_20ROI} for results for $R=20$ for RS data.

\textbf{Experiment 1.1: Decision stability under matched samples.} \label{exp-dataleak}
We first evaluated probes on the same samples used for training, changed only in montage, so that any train-test difference is attributable to the montage alone. 
\autoref{fig:functional_robustness}-(b,c) demonstrate an N2N-probe performance (diagonal) approximating to 1 across models and datasets. Thus, the probes are able to memorize the samples.
For the N2M-probes (off-diagonal), we observe minimal drops for CBraMod and REVE-ROI, and pronounced drops for LUNA and REVE-A in both tasks. SingLEM differs between the tasks, showing performance degradation in the RS task only.
REVE-A shows the largest drops despite sharing an encoder with the most robust setting, REVE-ROI, which points to the readout rather than the representation: a learned global query attending over the available channel-patch tokens is itself montage-conditioned, whereas ROI averaging is montage-invariant by construction. 
Furthermore, the drops of LUNA and REVE-A are smaller for RS than for MI, consistent with MI relying on a spatially focal set of sensorimotor channels \citep{Pfurtscheller_2001} that is more vulnerable to channel removal than the diffusely distributed eyes-open/closed signal.
LUNA and REVE-A are especially affected by these task-specifics, because their focal representation is redistributed across latent dimensions. This is in contrast to channel- and ROI-indexed readouts.


\textbf{Experiment 1.2: Cross-montage generalization to held-out data.} \label{exp-montage-gen}
To measure cross-montage generalization to new subjects for RS or epochs for MI, we train N2M-probes with subject- and epoch-level data splitting, respectively. Here, the performance difference between the N2N- and N2M-probes reflects the model’s ability to generalize across channel counts in unseen data. AUROC is considerably lower for MI than for RS (\autoref{fig:functional_robustness}-(d, e)), and MI performance varies strongly
across subjects (0.11--0.14 AUROC, \autoref{app:sec:numerical_results}). For RS, no consistent pattern of cross-montage generalization emerges: CBraMod, LUNA, and REVE-ROI perform best overall, with CBraMod and REVE-ROI showing the smallest montage-related drops, and SingLEM appearing robust but at a substantially lower baseline. REVE-A again performs worst and is the only setting reproducing the degradation pattern of Experiment 1.1. The MI results follow the same ordering, with a more pronounced advantage for REVE-ROI, but here CBraMod, LUNA, and REVE-A all lose AUROC under montage change, most strongly REVE-A and least strongly CBraMod.

\subsection{Experiment 2: Representational similarity}
\begin{figure}[t]
    \centering
    \includegraphics[width=0.8\linewidth]{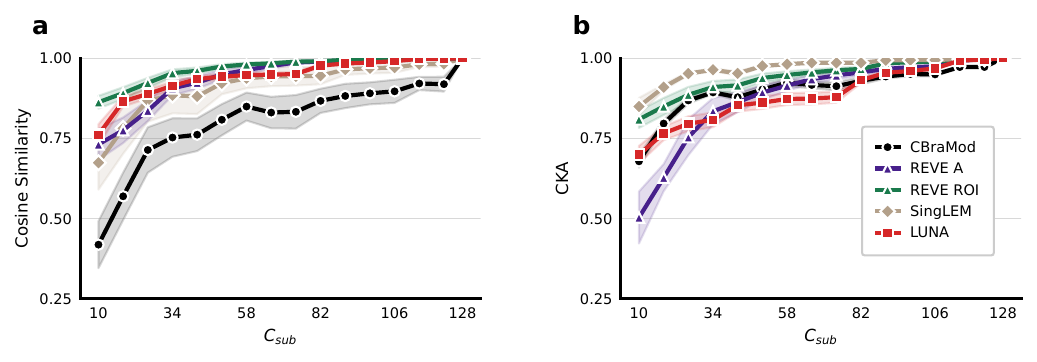}
    \caption{\textbf{Representational cross-montage robustness for resting-state EEG.} Each point compares the embedding of the full montage ($C_\text{full}$) to that of a subsampled montage with $C_\text{sub}$ channels, displayed on the x-axis. For within-subject similarities, each point is an average over subjects; variance is displayed over the first three EEG epochs per label (six trials per subject). \textbf{(a)} Within-subject cosine similarity. \textbf{(b)} Centered Kernel Alignment (CKA) between the subject-to-feature matrices of the two montages.}
    \label{fig:representational_robustness}
\end{figure}
\textbf{Protocol.}
We compare the embedding of the full recording montage $E^{C_\text{full}}$ to that of a spatially informed subsampled montage, $E^{C_\text{sub}}$, assessing within-subject consistency (cosine similarity; for Wasserstein distance, see \autoref{app:sec:models_embedding}) and across-subject structure (CKA \citep{kornblith2019cka}).
Both are compared using \textit{roi-concat-all} as the pooling for cosine similarity and CKA, except for LUNA, which bypasses $\phi$ and is compared with \textit{concat-all}. For each subject we recompute the embeddings for three EEG epochs per label across all subsampled montages. \autoref{fig:montage-sampling} illustrates some of the subsampled channel sets.

\textbf{Experiment 2.1: Within-subject representation similarity.}
The cosine similarity quantifies how an EEG embedding of a subject changes in comparison to the original montage as the channels are subsampled. 
A high cosine similarity suggests that the overall direction of the embedding vector changes little under the channel perturbation. Similarities rise towards 1 as $C_\text{sub}$ increases (\autoref{fig:representational_robustness}-(a), starting around 0.75 for all FMs except CBraMod, which starts near 0.5. The MI data show the same pattern with an even larger gap for CBraMod (\autoref{fig:representational_robustness_mi}). ROI-level cosine similarities and Wasserstein distances give a consistent picture (\autoref{fig:representational_robustness_roimean_wasserstein}).

\textbf{Experiment 2.2: Between-subject alignment.} 
CKA quantifies how much the pairwise structure between subject embeddings changes under subsampling. Stacking the flattened $\phi(E^{C_\text{full}})$ and $\phi(E^{C_\text{sub}})$ of $K$ subjects gives $X_\text{full}, X_\text{sub} \in \mathbb{R}^{K \times RPF}$, on which we evaluate $\texttt{CKA}(X_\text{full}, X_\text{sub})$. \autoref{fig:representational_robustness}-(b) exhibits trajectories of CKA as the channels are subsampled: the cross-subject structure becomes more aligned as $C_\text{sub}$ approaches $C_\text{full}$. The differences between FMs are small; only REVE-A embeddings are less aligned up to approximately $C_\text{sub} = 50$ channels, where it matches the other models. This illustrates an overall preservation of subject-specific differences in FM embeddings during montage perturbation. \autoref{fig:representational_robustness_mi}-(b) shows the resulting curves for MI data.


\section{Discussion and Conclusion}
\label{discussion}

Our results show that montage robustness is not a single property of an EEG foundation model. 
Stability of individual embeddings, preservation of between-subject geometry, and transfer of downstream decisions did not produce a consistent model ranking. 

For instance, representations changed in CBraMod under channel reduction without corresponding large performance losses; LUNA kept high within-instance similarity but lost downstream performance across settings; SingLEM varied by task and evaluation regime at a lower baseline, and REVE depended strongly on its readout.
These differences indicate that representational similarity alone cannot predict whether a downstream model will remain functional under montage change.

The contrast between REVE’s readouts provides the clearest evidence that robustness also depends on aggregation. ROI-pooled channel-wise embeddings showed smaller functional changes than the pretrained global attention readout, despite both originating from the same encoder.
ROI-pooling provides anatomically aligned feature positions across channel counts, whereas the global readout recomputes a summary from the channel-patch tokens available in each montage. This may make the global representation more sensitive to changes in channel availability.
However, ROI pooling is stabilizing by construction, so the result should not be read as a property of the REVE encoder but as evidence that anatomically aligned aggregation is a useful interface between variable-channel representations and downstream predictors.

We note some limitations for our study. The cross-model patterns are consistent with architectural hypotheses but do not identify their causes. The high within-instance similarity observed for LUNA and REVE may reflect their stronger integration of information across channels. Yet, while CBraMod processes cross-channel interactions, as well, it exhibits substantially lower similarity.
The evaluated models further differ in pretraining data, positional encoding, model size, feature dimensionality, and other design choices. Moreover, we tested each model with preprocessing matched to its pretraining configuration, mitigating distribution shifts. However, this limits the attribution of differences between models to their architecture alone. Model differences therefore characterize the complete pretrained systems rather than isolating the effect of any single channel-fusion mechanism.

Taken together, EEG-FMs that accept variable channel configurations at the input do not guarantee that representations or downstream predictions remain stable when montage/channel availability changes between training and deployment.
Evaluations should therefore examine both axes and treat the aggregation method as part of the robustness pipeline: robust deployment may require not only encoders that accept variable electrode sets, but readouts that preserve stable, task-relevant feature organization across them.

\begin{ack}
Justus Meyer zu Bexten is supported the Federal Ministry of Research, Technology and Space of Germany and by Sächsische Staatsministerium für Wissenschaft, Kultur und Tourismus in the programme Center of Excellence for AI-research „Center for Scalable Data Analytics and Artificial Intelligence Dresden/Leipzig“, project identification number: ScaDS.AI.
Laure Ciernik is funded by the Hector Fellow Academy and is grateful for their support.
\end{ack}

\section*{Author Contributions}
\textbf{Jakob Steglich}: Methodology, Software, Validation, Formal analysis, Investigation, Data Curation, Writing - Original Draft, Visualization, Project administration
\textbf{Justus Meyer zu Bexten}: Methodology, Writing - Original Draft, Software, Investigation
\textbf{Shakiba Moradi}: Methodology, Writing - Original Draft, Investigation
\textbf{Laure Ciernik}: Investigation
\textbf{Simon M. Hofmann}: Methodology, Validation, Investigation, Supervision, Project administration
\textbf{Mina Jamshidi Idaji}: Conceptualization, Methodology, Validation, Visualization, Investigation, Writing - Original Draft, Supervision, Project administration
\textbf{All authors}: Writing - Review \& Editing

\bibliographystyle{plainnat}
\bibliography{references}

\newpage

\appendix

\setcounter{figure}{0}
\renewcommand{\thefigure}{S\arabic{figure}}

\setcounter{table}{0}
\renewcommand{\thetable}{S\arabic{table}}

\section{Related Work: EEG-FMs} \label{app:sec:EEG_FM_overview}

We selected four EEG foundation models---SingLEM \citep{sukhbaatarSingLEMSingleChannelLarge2025a}, CBraMod \citep{wangCBraModCrissCrossBrain2024a}, LUNA \citep{donerLUNAEfficientTopologyAgnostic2025}, and REVE \citep{ouahidiREVEFoundationModel2025}---to represent different approaches to channel processing and spatial encoding. These architectural differences are directly relevant to our investigation of representation stability under changes in electrode number and configuration. An overview of the selected models and their architectural categories is provided in \autoref{eeg_foundation_models}. All four models use masked autoencoding/reconstruction-based pretraining, allowing us to focus on differences in their architectural treatment of channel information.
In this study, we focus on differences in their architectural treatment of channel information.

\begin{table}[bht]
\centering
\caption{Overview of the evaluated EEG foundation models.}
\label{eeg_foundation_models}
\renewcommand{\arraystretch}{1.2}
\setlength{\tabcolsep}{2.5pt}

\begin{tabular}{@{}
  >{\raggedright\arraybackslash\bfseries}p{0.9in}
  >{\raggedright\arraybackslash}p{0.8in}
  >{\raggedright\arraybackslash}p{1.05in}
  >{\raggedright\arraybackslash}p{1.05in}
  >{\raggedright\arraybackslash}p{1.05in}@{}}

\toprule
&
\textbf{SingLEM} &
\textbf{CBraMod} &
\textbf{LUNA} &
\textbf{REVE} \\

\midrule

Category &
Channel-independent &
Spatial-temporal channel interaction &
Latent channel unification &
Continuous coordinate-aware \\

\midrule

Pretraining corpus &
71 datasets &
1 (TUEG) &
2 (TUEG + Siena) &
92 datasets \\

\midrule

Cross-channel interaction &
Shared weights across channels; no interaction &
Separate spatial and temporal self-attention &
Learned query cross-attention &
Cross-channel self-attention \\

\midrule

Spatial encoding &
None &
Spatio-temporal convolution &
3D electrode coordinates &
4D Fourier encoding of electrode and temporal coordinates \\

\midrule

Channel/layout handling &
Channel dimension retained &
Channel dimension retained &
Fixed-size latent representation &
Coordinate-aware channel-temporal representation \\

\midrule

Pretraining channel configurations &
17--256  &
19 &
20, 22 (TUEG)\newline 29 (Siena) &
3--129 \\

\midrule

Related models &
-- &
BENDR, 
BIOT, 
Neuro-GPT, 
LaBraM, 
EEGPT, 
NeuroLM, 
BrainWave, 
FoME, 
CSBrain, 
EEGMamba, 
CodeBrain, 
BrainRVQ \footnotemark&
LuMamba \citep{broustail_lumamba_2026}, REVE (global token) &
NeurIPT \citep{fang_neuript_2025}, 
SAMBA \citep{hong_samba_2025}, 
HEAR \citep{chen_hear_2025} \\

\bottomrule

\end{tabular}
\end{table}
\footnotetext{BENDR: \cite{kostas_bendr_2021}; 
BIOT: \cite{yang_biot_2023};
Neuro-GPT: \cite{cui_neuro-gpt_2024};
LaBraM: \cite{jiang_large_2024};
EEGPT: \cite{wang_eegpt_2024};
NeuroLM: \cite{jiang_neurolm_2024};
BrainWave: \cite{yuan_brainwave_2024};
FoME: \cite{shi_fome_2024};
CSBrain: \cite{zhou_csbrain_2025};
EEGMamba: \cite{wang_eegmamba_2025};
CodeBrain: \cite{ma_codebrain_2025};
BrainRVQ: \cite{cui_brainrvq_2026};}
The four models evaluated in this work provide complementary approaches along two dimensions that are central to our study: \textit{cross-channel interaction} and \textit{spatial encoding} (see \autoref{app:sec:evaluated_models}).

\section{Evaluated models and embeddings}\label{app:sec:models_embedding}

\subsection{Evaluated models} \label{app:sec:evaluated_models}

In the following, we provide details about the four FMs used in this work, summarized in \autoref{tab:model_overview}.

\textbf{CBraMod.}
The backbone of CBraMod \citep{wangCBraModCrissCrossBrain2024a} is a criss-cross-transformer, which implements two attention mechanisms, whose aim is to model spatial and temporal relationships among the EEG data separately and in parallel. Data are segmented into $P$ patches, resulting in a tensor $\in \mathbb{R}^{C \times P \times T_p}$. The pretraining objective is a masked signal reconstruction.
Furthermore, the model introduces asymmetric conditional positional encoding (ACPE) to model the electrode position. These rely on convolutional neural networks (CNNs) to learn the positional embeddings from the spatial-temporal position of an EEG patch, which should enable the model to adapt to electrode layouts dynamically. 
Note that for downstream tasks, the adaptation to other montages still depends on the montages the model has seen during pretraining. Also, although the EEG patches attend to each other in the time and space domains, the model does not generate an explicitly fused representation of the input signal, the output embedding \(E\) has the shape \(E \in \mathbb{R}^{C \times P \times F}\).  

\textbf{LUNA.}
LUNA \citep{donerLUNAEfficientTopologyAgnostic2025} utilizes masked signal reconstruction. Segmented EEG-patches learn a temporal and frequency embedding. Afterwards, a positional embedding encodes the 3D electrode coordinates via NeRF sinusoidal projections \citep{Mildenhall_2020}. At the heart of LUNA lays a channel unification module, which learns fixed-sized query tokens \(Q\) that cross-attend to patch features. As a result, during channel unification, the channel count \(C\) is lost, and 2D output embeddings of shape \(E \in \mathbb{R}^{(Q \cdot P) \times F}\) are retained. These latent vectors are afterwards passed through multiple Transformer encoder blocks to learn temporal dependencies across patches before retrieving the final embedding.

\textbf{SingLEM.}
SingLEM \citep{sukhbaatarSingLEMSingleChannelLarge2025a} adopts an masked autoencoder MAE framework with signal reconstruction as the pretraining objective. In contrast to CBraMod and LUNA, the model does not implement any attention or learning of inter-channel dependencies. Instead, each channel is passed through the model independently. The patched EEG signal of a channel is forwarded through a 1D CNN to obtain a temporal embedding. The subsequent feature embedding module aims to represent short-range temporal relationships in contextualized embeddings. It applies multi-head self-attention on local context sequences for each token. Long-range temporal dependencies are afterwards captured by multi-head self-attention on the stacked contextualized embeddings. The final output embeddings are of shape \(E \in \mathbb{R}^{C \times P \times F}\). Crucially, SingLEM uses a late-fusion strategy for downstream applications, meaning the single-channel representations are concatenated for classification.

\textbf{REVE.}
REVE \citep{ouahidiREVEFoundationModel2025} uses a masked-autoencoder paradigm designed to explicitly generalize across varying electrode montages. It creates overlapping temporal EEG patches, projects them into linear patch embeddings, and then adds spatial information.
A 4D positional encoding based on the 3D electrode coordinates and temporal patch indices captures spatial information with Fourier features. These 4D encodings provide an adaptive function, which dynamically processes different electrode layouts. All positionally aware tokens are then processed jointly by a standard transformer encoder, where self-attention performs cross-channel fusion over coordinates and sequence length. The output embedding has a shape of \(E \in \mathbb{R}^{C \times P \times F}\). Simultaneously, during pretraining, an attention pooling is applied over the outputs of all Multi-Head Attention layers, which yields a global learned query token. We refer to this readout as \textit{REVE-A}. This token is used in REVE's architecture for a secondary reconstruction, which primarily exists to yield a more generalizable representation. This global query token can serve as a readout of the FM for collapsing the 3D-embeddings of the encoder into a fixed-size 1D representation for downstream classification tasks.

\begin{table}[t]
\centering
\small
\caption{\textbf{Overview of channel-fusion mechanisms across evaluated EEG foundation models.} All four models are pretrained with a masked-reconstruction objective; $C$, $P$, $F$, $Q$ follow the notation introduced in \autoref{app:sec:embedding_retrieval}.}
\label{tab:model_overview}
\begin{tabularx}{\textwidth}{@{} l >{\raggedright\arraybackslash}X >{\raggedright\arraybackslash}X c c @{}}
\toprule
\textbf{Model} & \textbf{Channel-Fusion Mechanism} & \textbf{Positional / Spatial Encoding} & \textbf{Output Shape} \\
\midrule
CBraMod & Criss-cross transformer: separate spatial and temporal self-attention branches & Asymmetric Conditional Positional Encoding (ACPE) & $C \times P \times F$ \\
\addlinespace
LUNA & Cross-attention: $Q \ll C$ learned latent queries attend to channel-wise patch features, collapsing the channel axis & 3D electrode coordinates via NeRF-style sinusoidal projection & $(Q \cdot P) \times F$ \\
\addlinespace
SingLEM & None during encoding. Channels processed fully independently (shared weights); cross-channel combination deferred to late-fusion concatenation downstream & (channel-independent) & $C \times P \times F$ \\
\addlinespace
REVE & Full self-attention over the flattened $(C \times P)$ token sequence + one summary token (cf. LUNA) & 4D Fourier features + learned linear projection (3D coords + temporal patch index) & $C \times P \times F$ \\
\bottomrule
\end{tabularx}
\end{table}

\subsection{Embedding retrieval} \label{app:sec:embedding_retrieval}
Let $k$ index subjects and $e$ index the samples collected per subject. A single preprocessed EEG sample is denoted as \(S_{k,e} \in \mathbb{R}^{C_{\text{full}} \times T}\), where $C_{\text{full}}$ is the number of channels and \(T\) is the number of time points in the trial. Each multi-channel signal is divided into $P$ patches of length $T_p$, which yields \(X_{k,e} \in \mathbb{R}^{C_{\text{full}} \times P \times T_p}\).
We additionally subsample the channel dimension from $C_{\text{full}}$ to $C_{\text{sub}}$ channels, yielding \(Y_{k,e} \in \mathbb{R}^{C_{\text{sub}} \times P \times T_p}\).
Next, both $\{X_{k,e}\}$ and $\{Y_{k,e}\}$ are passed through a pretrained foundation model $f(\cdot)$ with frozen weights to obtain embeddings
\begin{equation}
    E^{(\text{full}))}_{k,e} = f(X_{k,e}), \qquad E^{(\text{sub})}_{k,e} = f(Y_{k,e}).
\end{equation}
The shape of these embeddings depends on whether the model fuses the channel and patch dimension. CBraMod, SingLEM, and REVE preserve all dimensions, resulting in
\begin{equation}
    E^{(\text{full})}_{k,e} \in \mathbb{R}^{C_{\text{full}} \times P \times F}, \qquad E^{(\text{sub})}_{k,e} \in \mathbb{R}^{C_{\text{sub}} \times P \times F},
\end{equation}
where \(F\) is the embedding dimension per patch. 

LUNA and REVE-A fuse the channel dimension into $Q$ query tokens per patch, giving
\begin{equation}
    E^{(\text{full})}_{k,e} \in \mathbb{R}^{D \times F}, \qquad E^{(\text{sub})}_{k,e} \in \mathbb{R}^{D \times F},
\end{equation}
where \(D = Q \cdot P\) for LUNA and $D=1$ for REVE-A.

\paragraph{Regions of Interest (ROI)}
\begin{figure}[t]
    \centering
    \includegraphics[width=\linewidth]{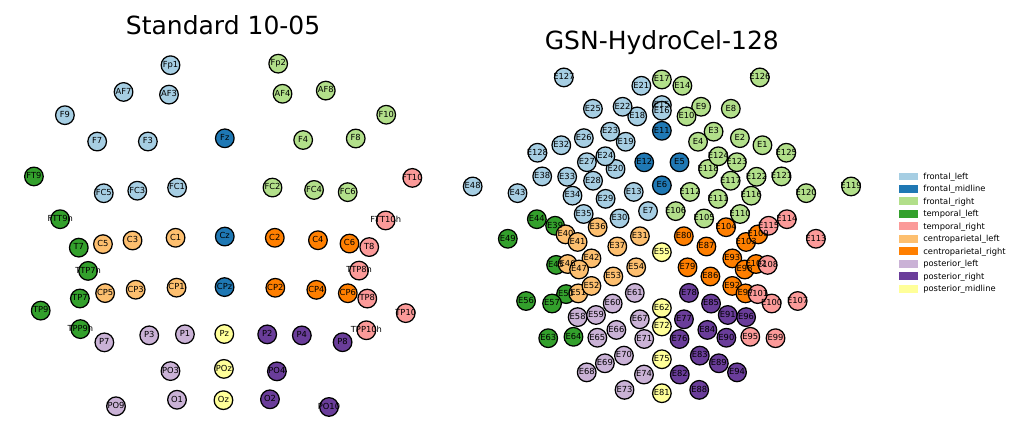}
    \caption{\textbf{Regions of Interest (ROIs) mapping for 128-channel GSN-HydroCel and 62-channel Standard-1005 montages.} These ROIs were used in experiments 1 and 2. Channels were assigned to ROIs manually based on their anatomical location.}
    \label{fig:montage-roi-mapping}
\end{figure}

\autoref{fig:montage-roi-mapping} shows how the Regions of Interest (ROIs) used for pooling are defined on the Standard-1005 montages used in the resting-state (RS) \citep{shirazi2024hbneeg} and motor imagery (MI) \citep{lee2019openbmi} datasets.
We decided to classify the ROIs based on neuroanatomical regions. For the Standard 10-05 montage, this could be achieved using the channel names, since these refer to such regions already. For the GSN montage, we categorized them based on our own qualitative assessment.

\section{Datasets and preprocessing details}\label{app:sec:dataset_prep_details}

\textbf{Resting-State (RS) Dataset.} 
The Healthy Brain Network EEG Dataset contains high-resolution EEG Data from over 3000 participants aged 5--21 years \citep{shirazi2024hbneeg}. The data are split across 11 releases. For our purposes, we utilize a random subsample with 100 participants. The signal has been downsampled to 100~Hz and bandpass-filtered between 0.5 and 50~Hz. The data have been recorded with a 129-channel GSN-HydroCel high-density layout.
Each participant went through ten 30-s epochs in which they were instructed to either open or close their eyes.
We selected the first 4~s of each 30-s trial for our downstream analyses.
For quality control, we clipped amplitudes at $500\,\mu\mathrm{V}$, and discarded epochs exceeding this threshold more than 5\% of the time. 

\textbf{Motor Imagery (MI) dataset.} 
The motor imagery dataset by \cite{lee2019openbmi} was recorded with a 62-electrode standard-1005 montage at a sampling rate of 1000~Hz from 52 subjects. Each trial followed a cued motor imagery paradigm: a 3-s preparatory fixation period, followed by a 4-s imagery task in which subjects imagined grasping with the hand indicated by a left or right visual cue.
The dataset comprises balanced left- and right-hand imagery trials. Trials from a training phase and an online test phase are available, although only the training-phase trials have labels and can be used for classification. Each subject underwent 100 trials in the train and test phases in two distinct recording sessions. Thus, for each of the 52 subjects, we obtained 200 labeled imagery trials.

\textbf{Preprocessing.} 
We applied minimal preprocessing steps, matching the pretraining configurations of the foundation models: average-referencing, resampling to 200 Hz for the CBraMod and REVE, 256 Hz for the LUNA, and 128 Hz for SingLEM followed by bandpass filtering in frequency bands 0.3--75.0, 0.1--75.0, 0.5--50, 0.5--99.5 for CBraMod, LUNA, SingLEM, and REVE, respectively. All bandpass filters were applied forward-backward to prevent phase shifts.

\textbf{Montage Subsampling.} 
For the experiments including linear probing, we subsample the 128-channel montage of the HBN resting-state dataset to both 64- and 32-channel montages, while the 62-channel recording montage of the MI dataset was subsampled to 32- and 16-channel layouts. For the MI dataset, the 32- and 16-channel montages are exact subsets of the 62-channel montage. In contrast, the reduced montages in the RS dataset differ slightly in electrode layout and do not have equivalent channel names. We therefore used nearest-neighbor electrode matching to arrive at layouts that closely resemble the native 64- and 32-channel GSN-HydroCel montages.

\begin{figure}[t]
    \centering
    \includegraphics[width=\linewidth]{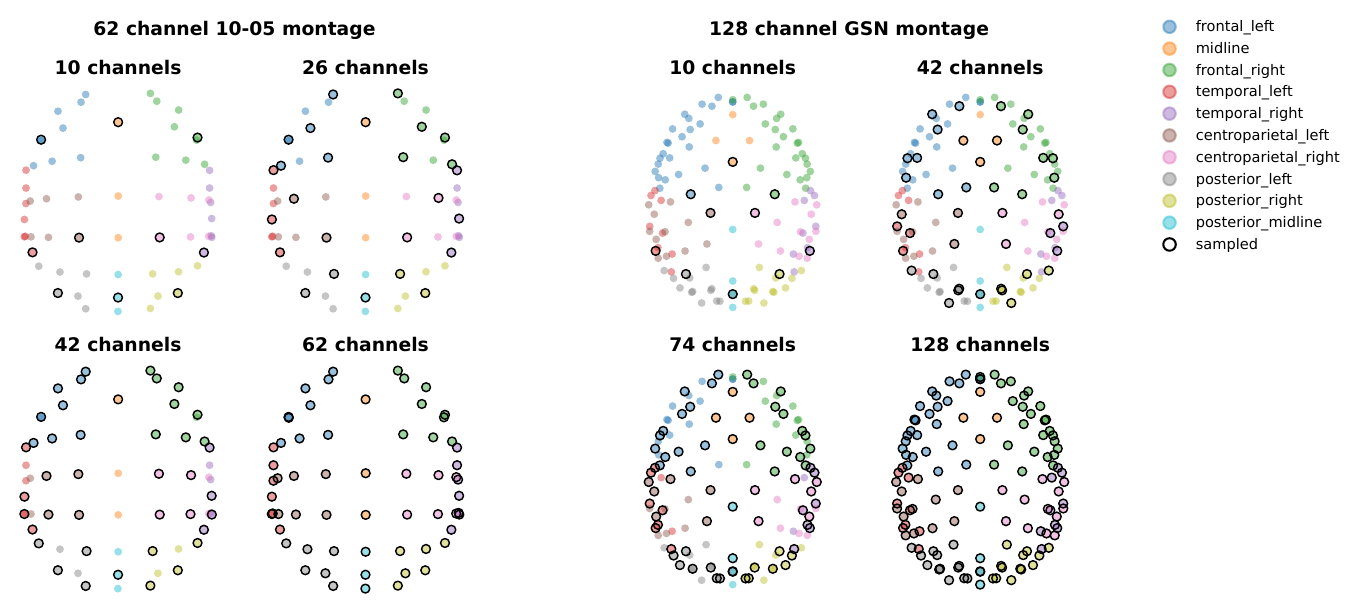}
    \caption{\textbf{Sampled montages for the 128-channel and 62-channel montages.}. These sampled channels were used in Experiment 2.}
    \label{fig:montage-sampling}
\end{figure}

For the representational space similarity experiments, we utilized a sampling procedure to determine the number of channels used for the construction of an embedding. This enables us to conduct those experiments for smaller steps of subsampled channels $C_{sub}$, where $C_{sub}$ progresses from 10 to 128 in steps of 10 and from 10 to 62 in steps of size 4, for the RS and MI datasets, respectively.
Specifically, we sample EEG channels while preserving spatial structure using farthest point sampling based on geodesic distance on one hemisphere, and mirror the selection to the other hemisphere. The midline channels are separately selected proportionally to the total amount of channels that were sampled. See \autoref{fig:montage-sampling} for examples of subsampled channel sets in the two datasets.

\section{Training details} \label{app:sec:model_training}
For both datasets, we apply a 2:1:1 train-validation-test split in Experiment 1.2, and repeat the training for three data splitting seeds, while the model stochasticity stays fixed. For cross-subject experiments the split is subject-level, and for the within-subjects experiments the split is on EEG epochs. 
Since both tasks are binary classification tasks, we implement the probes as logistic regression, and optimize only the \(C\) parameter on the validation set for regularization over a grid of [0.001, 0.01, 0.1, 1.0, 10.0, 100.0].
For evaluation, we report AUROC.


\section{Supplementary figures and tables} \label{app:sec:numerical_results}

\begin{table}[h]
    \caption{\textbf{AUROCs of within-montage baselines (N2N-probes with \textit{concat-all} pooling) in Experiment 1 for RS and MI data}. For the inter-subject RS task, the mean $\pm$ standard deviation of AUROC values across multiple seeds (n=10) is reported. For the intra-subject MI task, the values indicate the mean and standard deviation of AUROC values across subjects. Therefore, the high variance for MI task is associated with the high variance of subject-specific performances. This variation is in line with the CSP classification results reported in \cite{lee2019openbmi}.}
    \centering
    \small
    \setlength{\tabcolsep}{5pt}
    \begin{tabular}{lccc ccc}
        \toprule
        & \multicolumn{3}{c }{Resting-State task}
        & \multicolumn{3}{c}{Motor Imagery task} \\
        \cmidrule(lr){2-4} \cmidrule(l){5-7}
        Model
        & 32-ch & 64-ch & 128-ch
        & 16-ch & 32-ch & 62-ch \\
        \midrule
        CBraMod
        & 0.91 $\pm$ 0.02
        & 0.90 $\pm$ 0.01
        & 0.91 $\pm$ 0.01
        & 0.63 $\pm$ 0.12
        & 0.67 $\pm$ 0.13
        & 0.71 $\pm$ 0.14 \\
        
        LUNA
        & 0.90 $\pm$ 0.02
        & 0.89 $\pm$ 0.02
        & 0.91 $\pm$ 0.02
        & 0.60 $\pm$ 0.12
        & 0.61 $\pm$ 0.12
        & 0.62 $\pm$ 0.11 \\
        
        REVE
        & 0.94 $\pm$ 0.01
        & 0.92 $\pm$ 0.02
        & 0.93 $\pm$ 0.01
        & 0.69 $\pm$ 0.13
        & 0.69 $\pm$ 0.13
        & 0.69 $\pm$ 0.13 \\
        
        SingLEM
        & 0.89 $\pm$ 0.03
        & 0.88 $\pm$ 0.02
        & 0.91 $\pm$ 0.02
        & 0.62 $\pm$ 0.12
        & 0.65 $\pm$ 0.12
        & 0.67 $\pm$ 0.13 \\
        \bottomrule
    \end{tabular}
    \label{tab:N2N-prob-perf}
\end{table}

\begin{table}[h]
    \caption{\textbf{AUROCs of within-montage baselines (N2N-probes) in Experiment 1 by pooling method for RS and MI datasets}. Here, results for the full montages (128-channel for RS, and 62-channel for MI) are reported. For the inter-subject RS task, the mean $\pm$ standard deviation of AUROC values across multiple seeds (n=10) is noted. For the intra-subject MI task, the numbers indicate the mean and standard deviation of AUROC values across subjects. Therefore, the high variance for MI task is associated with the high variance of subject-specific performances. Additionally to the pooling methods introduced above, we report results for \textit{ch-mean} pooling, where the embeddings are averaged over the channel dimension, and concatenate patches and features, yielding 1D embeddings of size $(C \cdot P \cdot F)$. This method serves as baseline, since it removes most information across channels, while providing montage-agnostic embeddings.}
    \centering
    \small
    \setlength{\tabcolsep}{5pt}
    \begin{tabular}{lcccc}
        \toprule
        Method & concat-all & ROI-concat-all & ch-mean & attention-pool \\
        \midrule
        \multicolumn{5}{l}{\textit{MI task -- 62-channel}} \\
        \midrule
        CBraMod & 0.71 $\pm$ 0.14 & 0.70 $\pm$ 0.13 & 0.54 $\pm$ 0.08 & - \\
        LUNA    & 0.62 $\pm$ 0.11 & - & - & - \\
        REVE    & 0.69 $\pm$ 0.13 & 0.68 $\pm$ 0.13 & 0.56 $\pm$ 0.09 & 0.58 $\pm$ 0.09 \\
        SingLEM & 0.67 $\pm$ 0.13 & 0.67 $\pm$ 0.13 & 0.51 $\pm$ 0.05 & - \\
        \midrule
        \multicolumn{5}{l}{\textit{RS task -- 128-channel}} \\
        \midrule
        CBraMod & 0.910 $\pm$ 0.014 & 0.885 $\pm$ 0.016 & 0.790 $\pm$ 0.040 & - \\
        LUNA    & 0.911 $\pm$ 0.019 & - & - & - \\
        REVE    & 0.928 $\pm$ 0.013 & 0.906 $\pm$ 0.014 & 0.888 $\pm$ 0.023 & 0.790 $\pm$ 0.031 \\
        SingLEM & 0.912 $\pm$ 0.018 & 0.878 $\pm$ 0.023 & 0.608 $\pm$ 0.037 & - \\
        \bottomrule
    \end{tabular}
    \label{tab:pooling-combined}
\end{table}

\begin{figure}[t]
    \centering
    \includegraphics[width=0.9\linewidth]{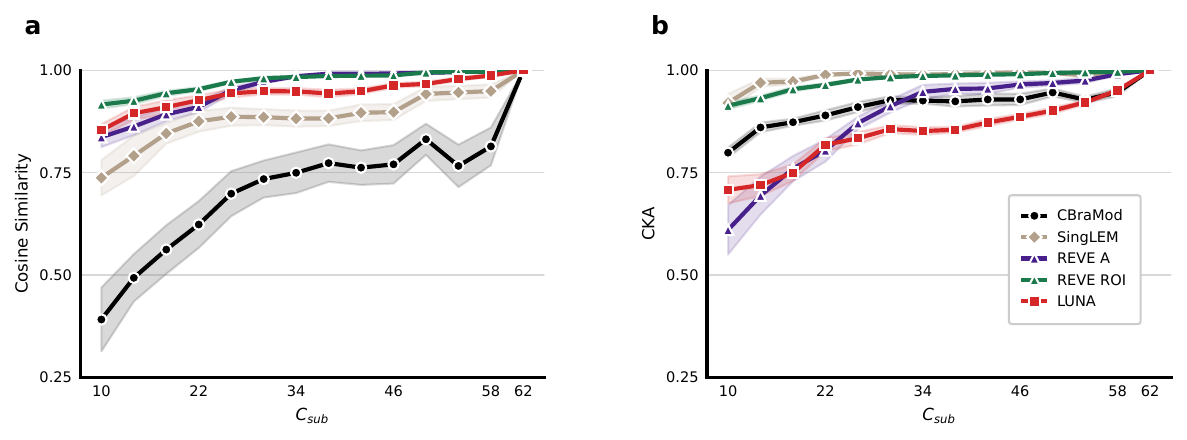}
    \caption{\textbf{Representational cross-montage robustness for motor imagery EEG.} Each point compares the embedding of the full montage ($C_\text{full}$) to that of a subsampled montage with $C_\text{sub}$ channels, displayed on the x-axis. For within-subject similarities, each point is an average over subjects; variance is displayed over the first three EEG trials per label (six trials per subject). \textbf{(a)} Within-subject cosine similarity. \textbf{(b)} Centered Kernel Alignment (CKA) between the subject-to-feature matrices of the two montages.}
    \label{fig:representational_robustness_mi}
\end{figure}

\begin{figure}[t]
    \centering
    \includegraphics[width=0.9\linewidth]{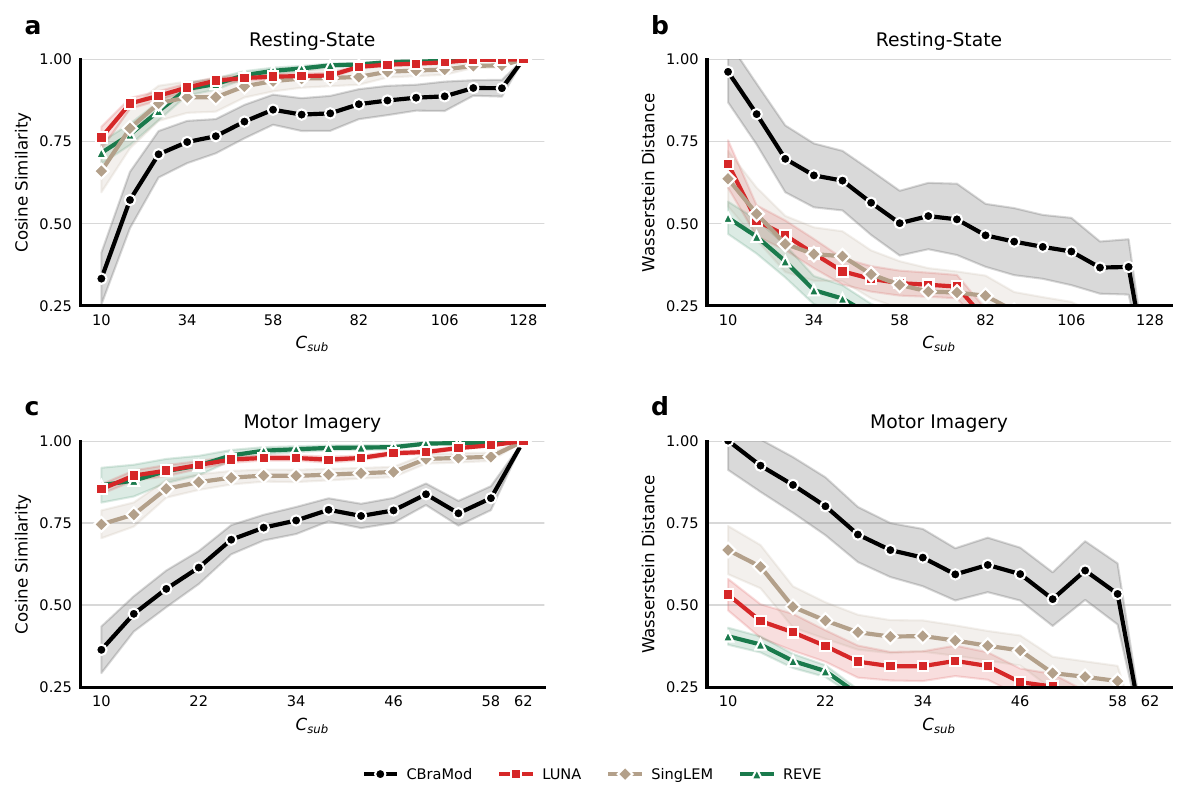}
    \caption{\textbf{Average cosine similarity across ROIs and Wasserstein Distances.} Each point compares the embedding of the full montage ($C_\text{full}$) to that of a subsampled montage with $C_\text{sub}$. Each point is an average over subjects, variance is displayed over the first three EEG trials per label (six trials per subject). \textbf{(a, c)} ROI-level within-subject cosine similarity: the ROI-pooled embeddings were first averaged over the patch dimension to produce $\in \mathbb{R}^{R \times F}$ embeddings. Then we computed $\left<s(f_r^{(1)}, f_r^{(2)})\right>_r$, where $f_r^{(i)}\in\mathbb{R}^{F}$ is the feature vector of ROI $r$ for montage $i$ and $s(.)$ is the cosine similarity. \textbf{(b, d)} Within-subject Wasserstein distances $W_2$ computed per subject between the two $(\mathrm{ROI}\cdot P) \times F$ embedding matrices, treating each of the $N = \mathrm{ROI}\cdot P$ rows as a sample in $\mathbb{R}^F$, except for LUNA, which has $N = \mathrm{Q}$ rows in $\mathbb{R}^F$.}
    \label{fig:representational_robustness_roimean_wasserstein}
\end{figure}


\begin{figure}[h]
    \centering
    \includegraphics[width=0.9\linewidth]{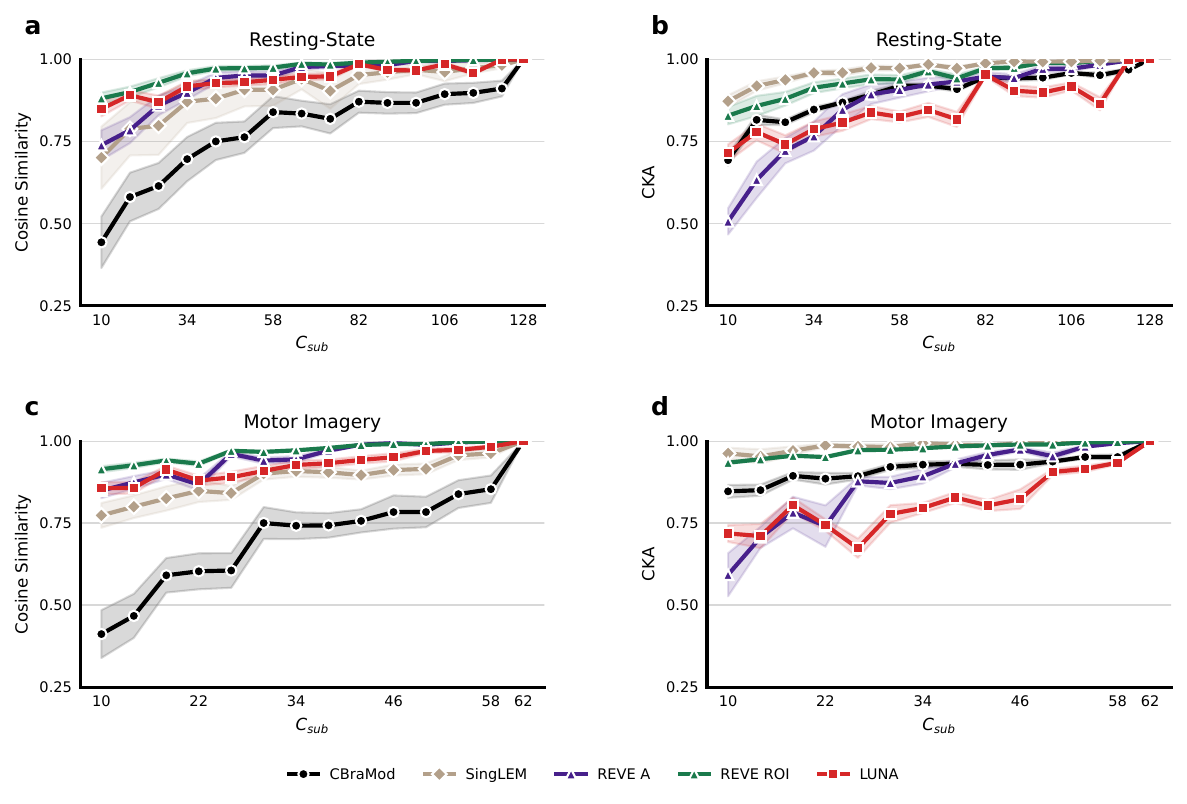}
    \caption{\textbf{Representational cross-montage robustness for resting-state and motor-imagery EEG with an alternative channel sampling seed.}  Each point compares the embedding of the full montage ($C_\text{full}$) to that of a subsampled montage with $C_\text{sub}$ channels, displayed on the x-axis. For within-subject similarities, each point is an average over subjects; variance is displayed over the first three EEG trials per label (six trials per subject). \textbf{(a, c)} Within-subject cosine similarity. \textbf{(b, d)} Centered Kernel Alignment (CKA) between the subject-to-feature matrices of the two montages.}
    \label{fig:representational_robustness_both}
\end{figure}

\begin{figure}[h]
    \centering
    \includegraphics[width=0.9\linewidth]{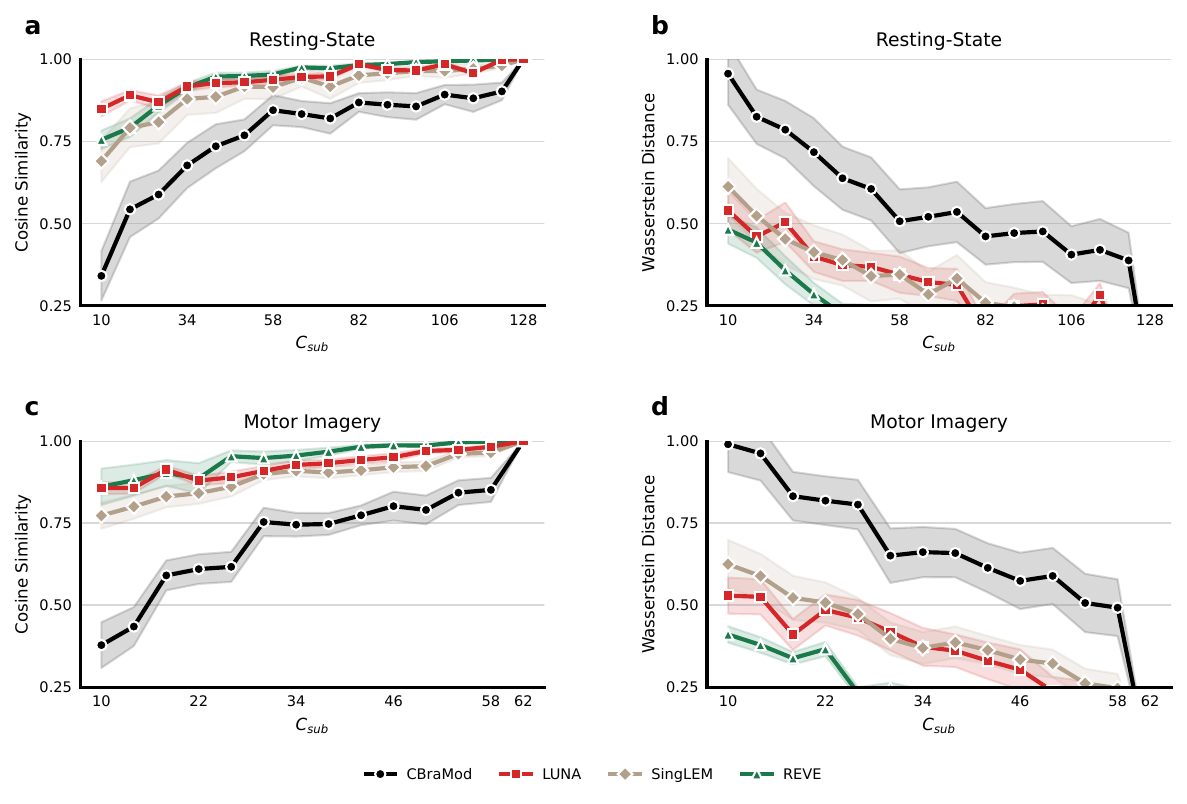}
    \caption{\textbf{Average cosine similarity across ROIs and Wasserstein Distances with an alternative channel sampling seed.} Each point compares the embedding of the full montage ($C_\text{full}$) to that of a subsampled montage with $C_\text{sub}$. Each point is an average over subjects, variance is displayed over the first three EEG trials per label (six trials per subject). \textbf{(a, c)} ROI-level within-subject cosine similarity: the ROI-pooled embeddings were first averaged over the patch dimension to produce $\in \mathbb{R}^{R \times F}$ embeddings. Then we computed $\left<s(f_r^{(1)}, f_r^{(2)})\right>_r$, where $f_r^{(i)}\in\mathbb{R}^{F}$ is the feature vector of ROI $r$ for montage $i$ and $s(.)$ is the cosine similarity. \textbf{(b, d)} Within-subject Wasserstein distances $W_2$ computed per subject between the two $(R\cdot P) \times F$ embedding matrices, treating each of the $N = R\cdot P$ rows as a sample in $\mathbb{R}^F$, except for LUNA, which has $N = \mathrm{Q}$ rows in $\mathbb{R}^F$.}
    \label{fig:representational_robustness_roimean_wasserstein_samp2}
\end{figure}


\begin{figure}[t]
    \centering
    \includegraphics[width=0.9\linewidth]{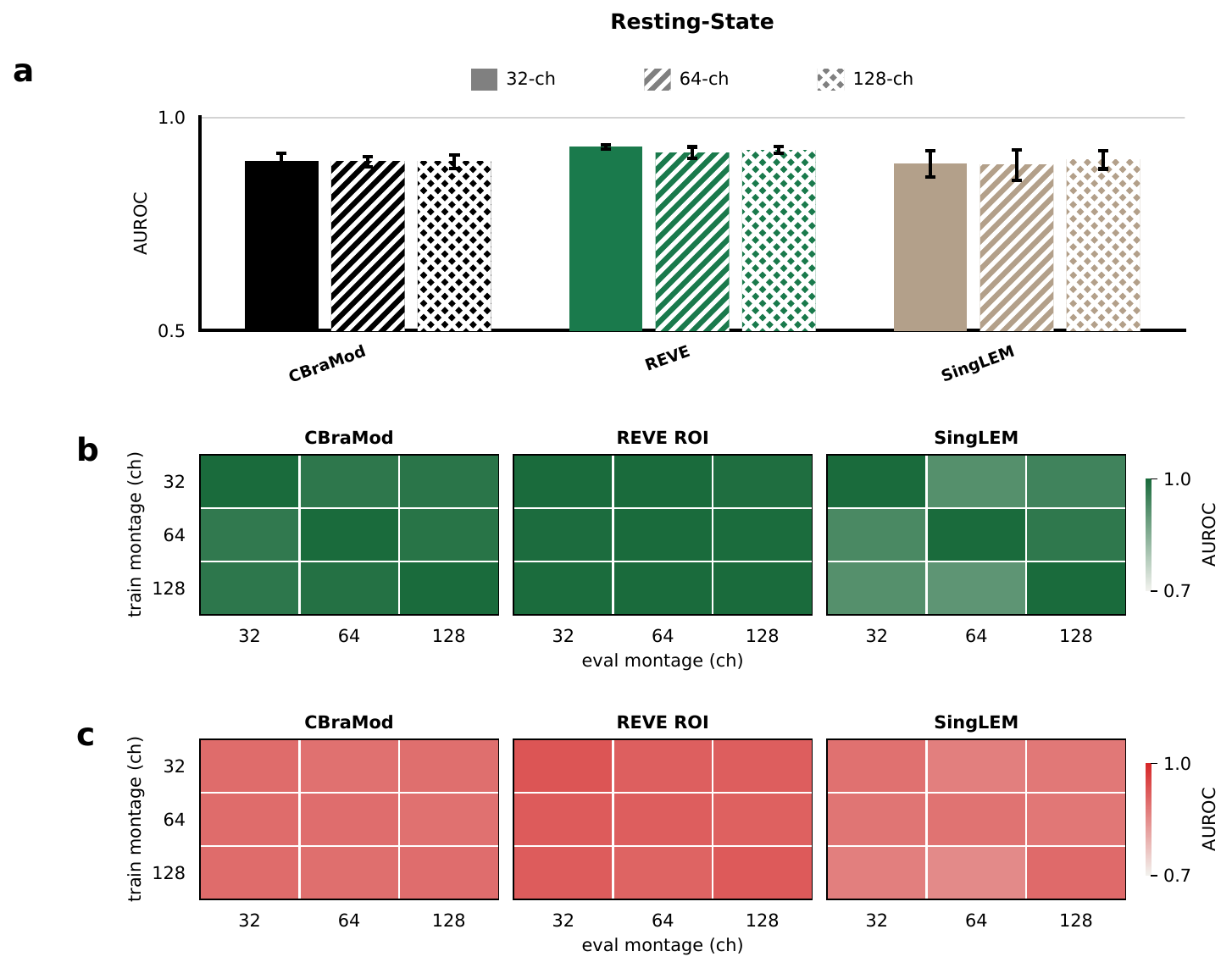}
    \caption{\textbf{Functional within- and cross-montage robustness for 20 ROIs.} \textbf{(a)} Within-montage baselines ($N = M$) for inter-subject RS classification; bars show mean AUROC and standard deviation across seeds for RS (\autoref{tab:N2N-prob-perf}). \textbf{(b)} N2M-probes on RS, with matched train and test samples, and \textbf{(c)} with held-out test data. Rows give the training and columns the test montage. The full montage has 128 channels for RS (\autoref{app:sec:dataset_prep_details}).}
    \label{fig:functional_robustness_20ROI}
\end{figure}

\end{document}